\documentclass[journal]{IEEEtran}

\ifCLASSINFOpdf
\else
   \usepackage[dvips]{graphicx}
\fi
\usepackage{url}
\usepackage{amsmath}
\usepackage{multirow}
\usepackage{amssymb}
\usepackage[capitalise]{cleveref}
\usepackage{graphicx}

\begin{document}

\title{HDBFormer: Efficient RGB-D Semantic Segmentation with A Heterogeneous Dual-Branch Framework}

\author{Shuobin Wei, Zhuang Zhou, Zhengan Lu, Zizhao Yuan, and Binghua Su }

\markboth{Journal of \LaTeX\ Class Files, Vol. 14, No. 8, August 2015}
{Shell \MakeLowercase{\textit{et al.}}: Bare Demo of IEEEtran.cls for IEEE Journals}
\maketitle

\begin{abstract}
In RGB-D semantic segmentation for indoor scenes, a key challenge is effectively integrating the rich color information from RGB images with the spatial distance information from depth images. However, most existing methods overlook the inherent differences in how RGB and depth images express information. Properly distinguishing the processing of RGB and depth images is essential to fully exploiting their unique and significant characteristics. To address this, we propose a novel heterogeneous dual-branch framework called HDBFormer, specifically designed to handle these modality differences. For RGB images, which contain rich detail, we employ both a basic and detail encoder to extract local and global features. For the simpler depth images, we propose LDFormer, a lightweight hierarchical encoder that efficiently extracts depth features with fewer parameters. Additionally, we introduce the Modality Information Interaction Module (MIIM), which combines transformers with large kernel convolutions to interact global and local information across modalities efficiently. Extensive experiments show that HDBFormer achieves state-of-the-art performance on the NYUDepthv2 and SUN-RGBD datasets. The code is available at: https://github.com/Weishuobin/HDBFormer.
\end{abstract}

\begin{IEEEkeywords}
RGB-D semantic segmentation, modality differences, heterogeneous dual-branch framework
\end{IEEEkeywords}
\IEEEpeerreviewmaketitle
\section{Introduction}

\begin{figure}
\centerline{\includegraphics[width=\columnwidth]{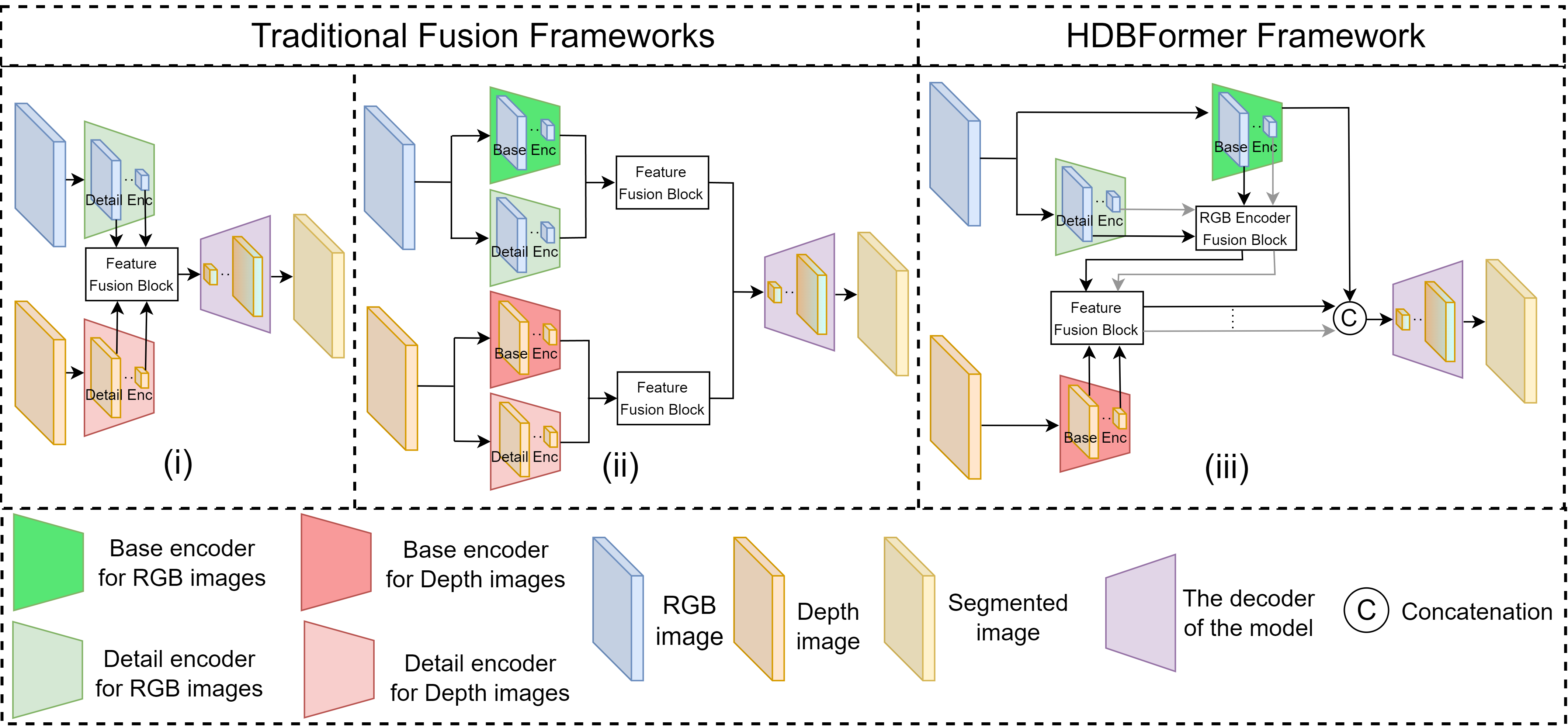}}
\caption{The difference between traditional fusion frameworks and HDBFormer fusion framework.}
\label{Fig:1}
\end{figure}

\IEEEPARstart{I}{ndoor}  scene segmentation is a key task in computer vision, aiming to categorize each pixel in an indoor image into its corresponding semantic category. Single RGB images often struggle with spatial relationships due to the lack of depth information \cite{noori2021survey}. To enhance segmentation accuracy, researchers have introduced combining RGB images with depth images (RGB-D) \cite{dong2023efficient}, \cite{wang2022multimodal}, \cite{yin2023dformer}, \cite{zhang2023cmx}, making the design of RGB-D modality fusion frameworks crucial. \cref{Fig:1} (left) shows two popular feature fusion frameworks, labeled (i) and (ii).

In framework (i), RGB and depth images are processed separately by hierarchical encoders and then fused \cite{zhang2021non}, \cite{zhang2024self}, \cite{hu2019acnet}, \cite{chen2020bi}, \cite{yue2021two}, \cite{zhang2024lightweight}. Zhang et al. \cite{zhang2021non} developed a non-local aggregation network (NANet) using a dual-branch hierarchical framework for efficient RGB-D feature fusion. In addition, Zhang et al. \cite{zhang2024self} combined Convolutional Neural Networks (CNN) and Transformers \cite{vaswani2017attention} to propose a dual-branch hierarchical feature extraction strategy to enhance feature extraction. However, these encoders may lack shallow features for precise object localization, which may lead to difficulties in accurately distinguishing objects in complex scenes. Framework (ii) addresses this with a dual-branch, two-stream cascade decoder that extracts both shallow and deep features by base and detail feature extraction, improving segmentation accuracy \cite{zhang2024dfti}, \cite{zhao2023cddfuse}, \cite{chen2024edge}, \cite{lee2023decomposed}. Zhang et al. \cite{zhang2024dfti} proposed a dual-branch, two-stream  fusion network combining Transformers and Inception, integrating an autoencoder and unsupervised learning for image fusion. Zhao et al. \cite{zhao2023cddfuse}  developed a relevance-driven fusion network using Restormer blocks and a dual-branch Transformer-CNN extractor for multi-modal feature processing.

Despite progress in RGB-D semantic segmentation, current research has overlooked that depth image features are less complex than RGB images. This oversight has led to two main issues in feature fusion design: 1) Most frameworks \cite{zhang2021non}, \cite{zhang2024self}, \cite{zhao2023cddfuse}, \cite{yue2021two}, \cite{yang2024min}, \cite{dong2022gebnet} use the same structure or degree of feature extraction for both RGB and depth images, failing to consider their modalities differences. 2) Using complex Transformers \cite{vaswani2017attention} or deep convolutional networks for depth image feature extraction often results in unnecessary computational costs and potential errors from overly complex feature extraction.

In this letter, we propose HDBFormer, a novel heterogeneous dual-branch framework for RGB-D semantic segmentation (\cref{Fig:2}). The framework uses different decoders depending on the image modality. For RGB images, a two-stream approach extracts shallow features for object localization and deep features for detailed segmentation, which are then merged by a specialised fusion module. For depth images, we design LDFormer, a hierarchical extractor that utilizes depthwise separable convolutions with downsampling block for efficient feature extraction. This approach significantly reduces parameters and computational complexity compared to using Transformers and deep convolutions for feature extraction from depth images \cite{zhang2024self}, \cite{hu2019acnet}, \cite{yue2021two}, \cite{li2023dctnet}, while still maintaining strong model expressiveness.

We also introduce the Modality Information Interaction Module (MIIM) to fuse RGB and depth image information, combining Transformers \cite{vaswani2017attention} and large kernel convolutions to enhance performance in complex indoor scene. MIIM is a well-designed feature fusion module that efficiently merges the rich color and texture information from RGB images with the spatial distance information from depth images. It introduces several key innovations: \textbf{1) Graded feature processing:} We classify features into primary and minor categories to prevent information overload and ensure effective fusion. \textbf{2) Optimized fusion strategy:} Unlike traditional methods that rely on convolution, element-wise addition and multiplication operations \cite{zhang2021non}, \cite{yue2021two}, \cite{dai2021attentional}, \cite{li2023dctnet}, \cite{yang2024min}, \cite{dong2022gebnet}, MIIM employs Transformers \cite{vaswani2017attention} for global feature extraction and large kernel convolution for precise local feature capture. This approach allows for more accurate modeling of complex relationships between image regions and enhances the model's ability to understand scene details. \textbf{3) Targeted fusion design:} Recognizing that RGB images contain rich color and texture information, while depth images provide simpler spatial information, we implemented a targeted fusion strategy to maximize the strengths of both modalities based on their inherent differences.

In summary, the main contributions of our work can be summarized as follows:

{•} We develop a novel and efficient heterogeneous dual-branch RGB-D semantic segmentation framework, which is designed to efficiently integrate multimodal information in the same scene. In particular, for depth images, we introduce LDFormer, a lightweight hierarchical encoder that efficiently extracts features with fewer parameters.

{•} We develop an innovative Modal Information Interaction Module (MIIM), which also employs a targeted fusion strategy designed to address the complexity differences between RGB and depth features. It also employs graded feature processing to prevent overloading and combines Transformers with large kernel convolution for efficient global and local information interaction.

{•} Extensive experiments demonstrate that HDBFormer achieves state-of-the-art performance on both the NYUDepthv2 and SUN-RGBD datasets.

\section{methodology}


\subsection{Encoder Module}
\noindent\textbf{1) RGB Image Feature Extraction.} For RGB image feature extraction, we use a two-stream framework. Deep features are extracted using a sophisticated  Swin Transformer \cite{liu2021swin} hierarchical feature extractor, while shallow features are extracted by a simple hierarchical encoder consisting of $3\times 3$ and $1\times 1$ convolutions with downsampling blocks. The four stages of features extracted by the Swin Transformer \cite{liu2021swin} are denoted as  $F_i^{Detail}$  (i = $1\sim4$), and those extracted by the simple encoder are denoted as  $F_i^{Base}$. Assuming that the current feature is \( F_i^{Base} \), the next feature \( F_{i+1}^{Base} \) is computed as:
\begin{equation}
F_{i+1}^{Base} = {MaxPool}({Conv}_{1 \times 1}({Conv}_{3 \times 3}(F_i^{Base}))).
\end{equation}
Here, $MaxPool(.)$ represents the max pooling operation. \( {Conv}_{1 \times 1} (.)\) and \( {Conv}_{3 \times 3} (.)\) denote the use of $1\times 1$ and $3\times 3$ convolutions, respectively.  These features are then integrated pairwise using the following formula:
\begin{equation}
\begin{split}
F_i^{RGB} = &\ {Conv}_{1 \times 1}({Concat}((F_i^{Depth} \oplus F_i^{Base}), \\
&\quad (F_i^{Depth} \otimes F_i^{Base}))).
\end{split}
\end{equation}
Here, $Concat(.)$ denotes concatenating along the channel dimension. \(\oplus \) and \(\otimes \) denote element-wise addition and element-wise multiplication, respectively. Element-wise multiplication is used to explore common features between channels, while element-wise addition is used to preserve as much detail as possible. With this fusion strategy, we obtain four levels of RGB features that fuse shallow and deep information.\vspace{10pt}

\begin{figure}
\centerline{\includegraphics[width=\columnwidth]{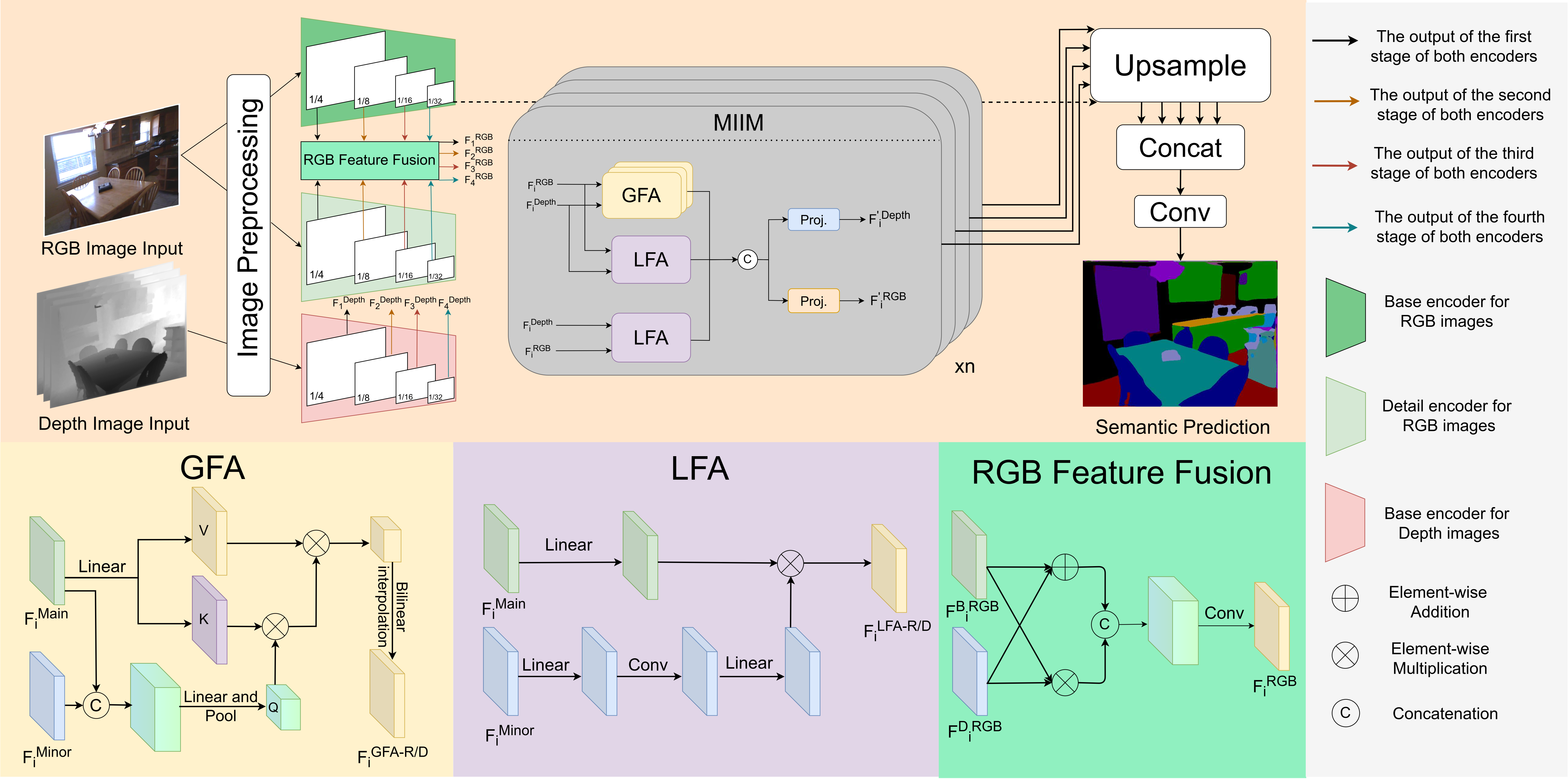}}
\caption{The network structure diagram of the HDBFormer.}
\label{Fig:2}
\end{figure}
\noindent\textbf{2) Depth Image Feature Extraction.} Depthwise separable convolutions have shown effectiveness in local feature extraction in early models \cite{howard2017mobilenets}, \cite{chollet2017xception}. Later, CAFormer \cite{yu2023metaformer} applied a Transformer-like hierarchical architecture for RGB image feature extraction. Building on these advancements, we develop LDFormer, using a hierarchical structure with depthwise separable convolutions and downsampling blocks for multi-level feature extraction. The primary function of the downsampling blocks is to match the feature dimensions extracted from depth images with those extracted from RGB images. Assuming that the current feature is represented as \( F_i^{Depth} \), the next feature \( F_{i+1}^{Depth} \) is calculated as:
\begin{equation}
F_{i+1}^{Depth}={MaxPool}({PWConv}_{1 \times 1}({DWConv}_{3 \times 3}(F_i^{Depth}))).
\end{equation}
Here, \( {DWConv}_{3 \times 3}(.) \) denotes the $3 \times 3$ depth convolution, and \( {PWConv}_{1 \times 1}(.) \) denotes $1 \times 1$ point-by-point convolution. 

Through the above operations, we extract four pairs of feature maps at different scales from RGB and depth images, with sizes \(\left[\frac{H}{4}, \frac{W}{4}, C\right]\), \(\left[\frac{H}{8}, \frac{W}{8}, 2C\right]\), \(\left[\frac{H}{16}, \frac{W}{16}, 4C\right]\), and \(\left[\frac{H}{32}, \frac{W}{32}, 8C\right]\). These feature maps are then paired and fused in the information interaction module using a specially designed method.

\subsection{Modality Information Interaction Module}

The feature fusion process of MIIM is systematic and efficient. Firstly, the features $F_i^{RGB}$ and $F_i^{Depth}$ extracted by the encoder are fed into the interaction module. To effectively fuse global and local information, we design two specialised sub-modules: the Global Fusion Attention (GFA) Module and the Local Fusion Attention (LFA) Module. GFA primarily fuses global information from RGB and depth images, while LFA focuses on merging local details and regional information.

In the fusion process, the input feature pairs are explicitly divided into primary and minor features, denoted as $F_i^{{Main}}$ and $F_i^{{Minor}}$, respectively. The primary features are used as the main role of fusion, while the minor features play a supporting and complementary role. If the RGB image is designated as the primary feature and the depth image as the minor feature, the feature pair will be processed by GFA and LFA to generate global and local information dominated by the RGB image. Conversely, if the depth image is used as the primary feature and the RGB image as the minor feature, the feature pair will undergo only LFA processing to generate local features dominated by depth information. \cref{Fig:2} (below) details the GFA and LFA modules and their operational flow.\vspace{10pt}

\noindent\textbf{GFA.}
The core design of the GFA module focuses on achieving global fusion of RGB and depth image features. Inspired by the Transformer’s effectiveness in multimodal integration \cite{wang2022rtformer}, \cite{zhang2023cmx}, the GFA module utilizes the Transformer structure for robust global feature processing. Additionally, the query ($Q$) is pooled to reduce the computational costs required for dot product calculations \cite{wang2022rtformer}. To compute the query ($Q$), we first concatenate the primary and minor features,  $F_i^{{Main}}$  and  $F_i^{{Minor}}$ , and then downsample them using a $6\times6$ pooling layer. The key ($K$) and value ($V$) are separately computed by applying two linear transformations to  $F_i^{{Main}}$ . Thus, the formulas for computing $Q$, $K$, and $V$ in the GFA module are as follows:
\begin{equation}
\left\{
\begin{aligned}
Q &= {Linear}({Pool}_{6 \times 6}({Concat}(F_i^{{Main}}, F_i^{{Minor}}))), \\
K &= {Linear}(F_i^{{Main}}), \\
V &= {Linear}(F_i^{{Main}}),
\end{aligned}
\right.
\end{equation}
where $Linear(.) $ denotes the application of a linear transformation layer. $Pool_{6 \times 6}(.) $ denotes the reduction of the feature size to $6\times 6$. Where  \(Q\) $\in$ \(R^ {6 \times 6 \times C^d} \), \(K\) and \(V\) $\in$ \(R^ {H \times W \times C^d} \). Using these features for dot product operation, we can get the formula of GFA:
\begin{equation}
\label{eq. 5}
F_i^{GFA-R/D} = {BI}_{H \times W}\left(V \cdot {Softmax}\left(\frac{Q^T \cdot K}{\sqrt{C^d}}\right)\right).
\end{equation}
In \cref{eq. 5}, if the primary features of the input are RGB images, the output is $F_i^{GFA-R}$ and if the primary features of the input are depth images, the output is $F_i^{GFA-D}$. In addition, ${BI}_{H \times W}(.)$ denotes the use of bilinear interpolation to resize the features from $6\times 6$ back to the original size $H\times W$.\vspace{10pt}

\noindent\textbf{LFA.} Our LFA enhances regional feature expression by using linear transformations and large-kernel convolution operations \cite{qin2024lkformer}, \cite{ding2022scaling}, \cite{li2024lkca}, enabling the model to precisely capture and integrate local features. This approach complements global features by improving the model’s sensitivity to localized information. Therefore, the arithmetic formula of LFA can be expressed as:
\begin{equation}
\begin{split}
F_i^{LFA-R/D} = &\ {Linear}(F_i^{{Main}}) \odot {Linear}\\
& ({Conv}_{7 \times 7}({Linear}(F_i^{{Minor}}))).
\end{split}
\end{equation}
After the above process, we obtain three features. Next, we concatenate these features and process them through two linear operations respectively to generate two new features that match the dimensions of the input features. This can be represented by the following formula:
\begin{equation}
\left\{
\begin{aligned}
F{'}_{i}^{RGB} &= {Linear}({Concat}(F_{i}^{GFA-R}, F_{i}^{LFA-R}, F_{i}^{LFA-D})), \\
F{'}_{i}^{Depth} &= {Linear}({Concat}(F_{i}^{GFA-R}, F_{i}^{LFA-R}, F_{i}^{LFA-D})).
\end{aligned}
\right.
\end{equation}
To further optimize information interaction between RGB and depth images, we adopt an iterative enhancement strategy. Specifically, we perform the information interaction process \(N\) times, with \(N\) set to 2. The iterative enhancement strategy progressively strengthens information exchange between RGB and depth features through multiple fusion steps, resulting in deeper and more robust feature integration. 
\subsection{Decoder Module}
In our decoder module, we first upsample the last feature in the base encoder of the RGB image as well as all the features obtained from the information interaction module to align them to the same feature size. Then, we concatenate these upsampled features and adjust the number of channels by a $1\times 1$ convolution block to match the number of categories in the prediction task. Thus, the operations in the decoder can be represented by the following formula:
\begin{equation}
\begin{split}
X_{{Output}} = {Conv}_{1 \times 1}({Concat}({UP}(F{'}_{1}^{RGB}, F{'}_{2}^{RGB}, \\
F{'}_{3}^{RGB}, F{'}_{4}^{RGB}, F_{4}^{{Base}}))),
\end{split}
\end{equation}
where \({UP(.)} \) denotes the upsample operation, which upsamples the dimensions of the five features to a uniform size for subsequent stitching operations.
\section{experiment}

\subsection{Dataset and Evaluation Metrics}
We select two popular RGB-D semantic segmentation datasets for testing: NYUDepthv2 \cite{silberman2012indoor} and SUN-RGBD \cite{song2015sun}. NYUDepthv2 includes 1449 RGB-D samples, with 795 for training and 654 for testing, covering 40 different categories; SUN-RGBD contains 10335 RGB-D samples, with 5285 for training and 5050 for testing, spanning 37 categories. To assess the performance of different methods, we use the commonly adopted metrics: mean Intersection over Union (mIoU) and Pixel Accuracy (Pixel Acc).

\begin{figure}
\centerline{\includegraphics[width=\columnwidth]{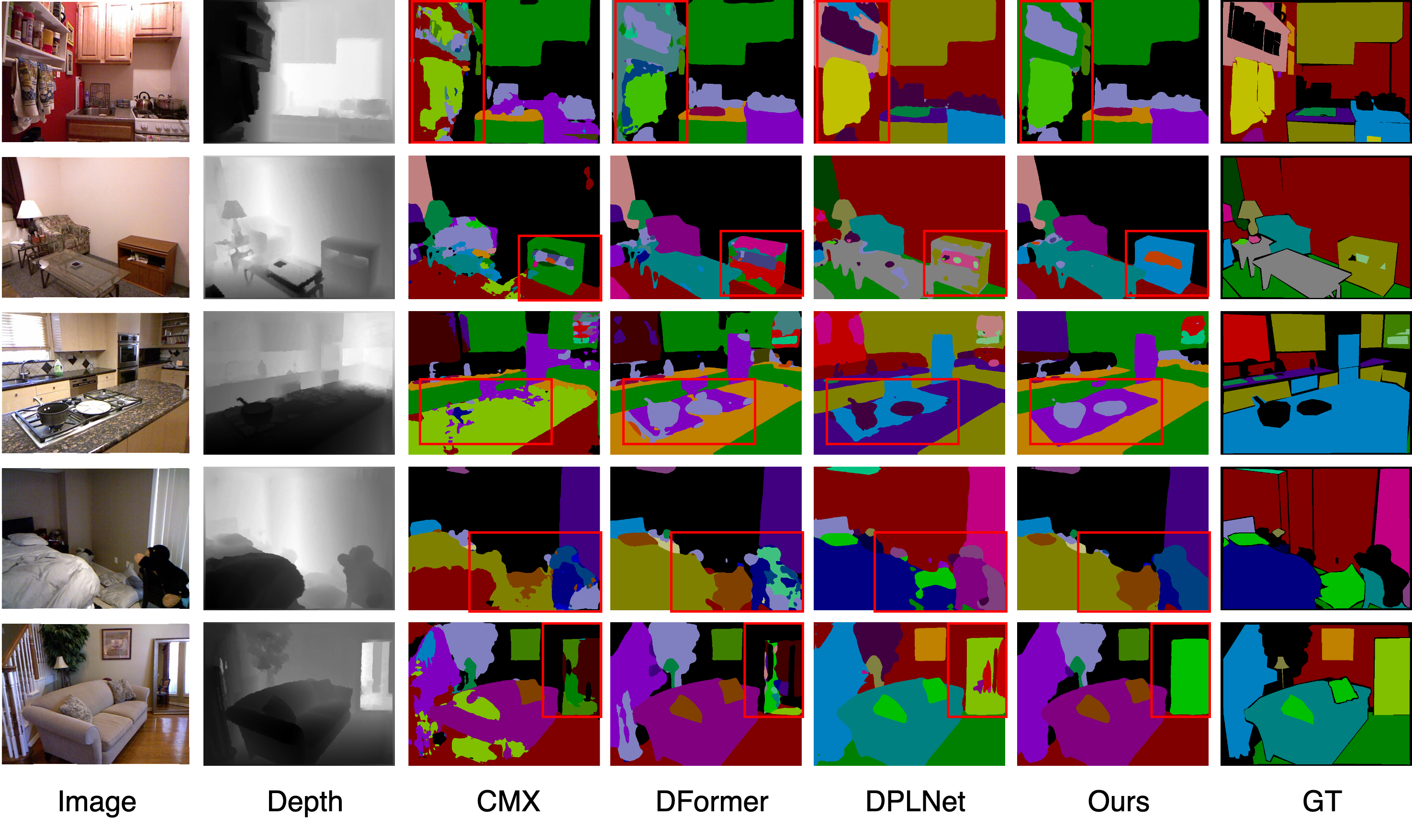}}
\caption{The segmentation results of HDBFormer on the NYUDepthv2 dataset are compared with CMX \cite{zhang2023cmx}, DFormer \cite{yin2023dformer} and DPLNet \cite{dong2023efficient}.}
\label{Fig:3}
\end{figure}
\begin{table}[]
\caption{Comparison of HDBFormer's performance with 15 other models in the NYUDepthv2 and SUN-RGBD datasets. The best results in the table are shown in bold.}
\label{Tab:1}
\setlength{\tabcolsep}{2.1pt}
\begin{tabular}{lcccccc}
\hline
\multicolumn{2}{c}{}                                       &                            & \multicolumn{2}{c}{NYUDepthv2} & \multicolumn{2}{c}{SUN-RGBD}     \\ \cline{4-7} 
\multicolumn{2}{c}{\multirow{-2}{*}{Model \& Publishment}} & \multirow{-2}{*}{Backbone} & PA            & mIoU           & PA             & mIoU           \\ \hline\hline
ACNet \cite{hu2019acnet}                                   & ICIP 2019       & ResNet-50                  & —             & 48.3           & —              & 48.1           \\
SA-Gate \cite{chen2020bi}                                 & ECCV 2020       & ResNet-101                 & 77.9          & 52.4           & 82.5           & 49.4           \\
PDCNet \cite{yang2023pixel}                                     & IEEE TCSVT    & ResNet-101                 & 78.4          & 53.5           & 83.3           & 49.6           \\
DCANet \cite{bai2022dcanet}                                     & —     & ResNet-101                 & 78.2          & 53.3           & 82.6           & 49.6           \\
TCD \cite{yue2021two}                                     & IEEE SPL        & ResNet-101                 & 77.8          & 53.1           & 83.1           & 49.5           \\
NANet \cite{zhang2021non}                                   & IEEE SPL        & ResNet-101                 & 77.9          & 52.3           & 82.3           & 48.8           \\
CEN \cite{wang2020deep}                                     & NeurIPS 2020    & ResNet-152                 & 77.4          & 52.5           & 83.2           & 51.1           \\
AsymFormer \cite{du2024asymformer}                                     & CVPR 2024      & MiT-B0                     & —           & 55.3           & 81.9              & 49.1           \\
CMNext \cite{zhang2023delivering}                                  & CVPR 2023       & MiT-B4                     & —             & 56.9           & —              & 51.9           \\
CMX \cite{zhang2023cmx}                                     & IEEE T-ITS      & MiT-B5                     & 80.1          & 56.9           & —              & 52.4           \\
ComPtr \cite{pang2023comptr}                                  & —               & Swin-B                     & 79.5          & 55.5           & —              & —              \\
OMNIVORE \cite{girdhar2022omnivore}   & CVPR 2022       & Swin-B                     & —             & 55.1           & —              & —              \\
LDSNet \cite{zhang2024lightweight}                                  & IEEE SPL        & LDSNet-base                & 79.4          & 56.3           & 82.7           & 51.8           \\
DFormer \cite{yin2023dformer}                                 & ICLR 2024       & DFormer-L                  & 80.2         & 57.2           & 83.8          & 52.5           \\
DPLNet \cite{dong2023efficient}                                 & IROS 2024       & MiT-B5                  & —         & \textbf{59.3}           & —          & 52.8           \\\hline
\multicolumn{2}{c}{HDBFormer (Ours)}                        & Swin-B                     & \textbf{81.0} & \textbf{59.3} & \textbf{84.2} & \textbf{53.9} \\ \hline
\end{tabular}
\end{table}

\subsection{Implementation Details}
All our experiments are conducted on a system equipped with an NVIDIA RTX4090 GPU. During training, we apply various data augmentation techniques to improve the model’s generalization ability, including random flipping and random scaling (ranging from 0.5 to 1.75). On the NYUDepthv2 and SUN-RGBD datasets, we set the batch size to 4, the initial learning rate to 1e-5, and the weight decay to 1e-2. To optimize the model, we use the cross-entropy loss function and the AdamW \cite{kingma2014adam} optimizer, and we implement a polynomial decay strategy for adjusting the learning rate. Additionally, we employ a multi-scale flip inference strategy, using scales of \{0.5, 0.75, 1, 1.25, 1.5\}.

\subsection{Comparison with State-of-the-Art Method}
To demonstrate the effectiveness of the proposed method, we compare HDBFormer with 15 other RGB-D semantic segmentation models. As shown in \cref{Tab:1}, HDBFormer based on Swin-B achieves 59.3\% of mIoU and 81.0\% of Pixel Acc on the NYUDepthv2 dataset, and achieves 53.9\% of mIoU and 84.2\% of Pixel Acc on the SUN-RGBD dataset. Comparing with other backbones (e.g. ResNet \cite{he2016deep}, MiT \cite{xie2021segformer} ), and even compared to models that also use Swin-B as the backbone, our method demonstrates state-of-the-art performance in several metrics. \cref{Fig:3} visualizes the segmentation effectiveness differences between HDBFormer and other models, including CMX \cite{zhang2023cmx}, DFormer \cite{yin2023dformer}, and DPLNet \cite{dong2023efficient}. It can be observed that HDBFormer more effectively utilizes spatial depth information from depth images, achieving more accurate segmentation results in RGB images with diverse and complex color details.

\begin{table}
\caption{Ablation experiment of LDFormer at NYUDepthv2 dataset.}
\label{Tab:2}
\setlength{\tabcolsep}{5pt}
\begin{tabular}{llccc}
\hline
Depth Image Encoder & Method      & mIoU          & FLOPs(G)     & Params(M) \\ \hline
Swin-T              & Transformer & 58.4          & 28.6         & 27.5              \\
Swin-B              & Transformer & 57.1          & 99.6         & 86.7              \\
MiT-B2              & Transformer & 58.2          & 19.2         & 25.4              \\
MiT-B4              & Transformer & 57.9          & 52.9         & 62.6              \\
Resnet-50           & CNN         & 59.0          & 25.3         & 11.7              \\
Resnet-152          & CNN         & 58.0          & 71.0         & 60.2              \\
Ordinary Conv       & CNN         & 58.0          & 14.1         & 6.3               \\ \hline
LDFormer (Ours)      & CNN         & \textbf{59.3} & \textbf{2.9} & \textbf{0.7}      \\ \hline
\end{tabular}
\end{table}

\begin{table}[]
\caption{Ablation experiment of MIIM feature fusion module on NYUDepthv2 dataset.}
\label{Tab:3}
\setlength{\tabcolsep}{16pt}
\begin{tabular}{llllc}
\hline
\multicolumn{4}{c}{The Element of MIIM} & \multirow{2}{*}{mIoU} \\ \cline{1-4}
LFA1      & GFA1      & GFA2          & LFA2     &                       \\ \hline
\checkmark         & \checkmark         & \checkmark             & \checkmark        & 58.3                    \\
\checkmark         & \checkmark         & \checkmark             &          & 57.9                  \\
\checkmark         &           & \checkmark             & \checkmark        & 57.8                  \\
          & \checkmark         & \checkmark             & \checkmark        & 58.1                  \\
\checkmark         & \checkmark         &               &          & 58.3                  \\
          & \checkmark         &               & \checkmark        & 57.6                  \\
\checkmark         &           & \textbf{}     & \checkmark        & 57.6                  \\ \hline
\checkmark         & \checkmark         &               & \checkmark        & \textbf{59.3}         \\ \hline
\end{tabular}
\end{table}

\subsection{Ablation Studies}
\noindent\textbf{1) Ablation experiments with depth image encoder.}
To evaluate the performance of different depth image encoders, we compare LDFormer with several mainstream encoders, as shown in \cref{Tab:2}. LDFormer achieves the best performance with minimal parameters and FLOPs, demonstrating its high efficiency. Notably, Transformer-based networks generally underperform compared to CNN-based networks, and more complex encoders often yield poorer results in depth feature extraction. This suggests that simpler encoders are sufficient for depth images with distinct features.\vspace{10pt}

\noindent\textbf{2) Ablation experiments with components of MIIM.}
We conduct a detailed evaluation of each component in MIIM. The GFA and LFA components focused on RGB features are labeled GFA1 and LFA1, while those focused on depth features are labeled GFA2 and LFA2. As shown in \cref{Tab:3}, by removing each component individually, we find that the model achieves the best performance when depth features bypass the GFA module. This suggests that for depth features of low complexity, excellent results can be obtained without the need for complex GFA with RGB features.

\section{Conclusion}
This letter proposes a novel solution for parsing complex RGB-D indoor scenes. Most RGB-D semantic segmentation models overlook the information differences between RGB and depth images. To address this, we design HDBFormer, a heterogeneous dual-branch architecture that adapts feature extraction complexity to the modality differences between RGB and depth images. Key components include LDFormer, a lightweight depth feature extractor, and the MIIM module, which fuses RGB and depth information using Transformers and large kernel convolutions. Extensive experiments have shown that HDBFormer achieves state-of-the-art results on both the NYUDepthv2 and SUN-RGBD datasets. In future work, we will further explore detailed feature extraction for depth images and continue optimizing the model structure to enable applications in other cross-modal fusion fields, such as RGB-T scene parsing and autonomous driving analysis.


\clearpage
\begin{thebibliography}{1}

\bibitem{noori2021survey} A. Y. Noori, "A survey of RGB-D image semantic segmentation by deep learning," in \textit{Proc. 2021 7th Int. Conf. Adv. Comput. Commun. Syst. (ICACCS)}, vol. 1, pp. 1953-1957, 2021.
\bibitem{zhang2021non} G. Zhang, J.-H. Xue, P. Xie, S. Yang, and G. Wang, "Non-local aggregation for RGB-D semantic segmentation," \textit{IEEE Signal Processing Letters}, vol. 28, pp. 658-662, 2021.
\bibitem{zhang2024self} H. Zhang, X. Ran, and W. Zhou, "Self-Knowledge Distillation-Based Staged Extraction and Multiview Collection Network for RGB-D Mirror Segmentation," \textit{IEEE Signal Processing Letters}, 2024.
\bibitem{zhang2024dfti} Z. Zhang, D. Zhou, G. Sun, Y. Hu, and R. Deng, "DFTI: Dual-branch Fusion Network based on Transformer and Inception for Space Non-cooperative Objects," \textit{IEEE Transactions on Instrumentation and Measurement}, 2024.
\bibitem{zhao2023cddfuse} Z. Zhao, H. Bai, J. Zhang, Y. Zhang, S. Xu, Z. Lin, R. Timofte, and L. Van Gool, "Cddfuse: Correlation-driven dual-branch feature decomposition for multi-modality image fusion," in \textit{Proc. IEEE/CVF Conf. Comput. Vis. Pattern Recognit.}, pp. 5906-5916, 2023.
\bibitem{dong2023efficient} S. Dong, Y. Feng, Q. Yang, Y. Huang, D. Liu, and H. Fan, "Efficient multimodal semantic segmentation via dual-prompt learning," \textit{Proceedings of the IEEE/RSJ International Conference on Intelligent Robots and Systems (IROS)}, 2024.
\bibitem{wang2022multimodal} Y. Wang, X. Chen, L. Cao, W. Huang, F. Sun, and Y. Wang, "Multimodal token fusion for vision transformers," in \textit{Proc. IEEE/CVF Conf. Comput. Vis. Pattern Recognit.}, pp. 12186-12195, 2022.
\bibitem{yin2023dformer} B. Yin, X. Zhang, Z. Li, L. Liu, M.-M. Cheng, and Q. Hou, "Dformer: Rethinking RGB-D representation learning for semantic segmentation," \textit{Proceedings of the International Conference on Learning Representations (ICLR)}, 2024.
\bibitem{zhang2023cmx} J. Zhang, H. Liu, K. Yang, X. Hu, R. Liu, and R. Stiefelhagen, "CMX: Cross-modal fusion for RGB-X semantic segmentation with transformers," \textit{IEEE Transactions on Intelligent Transportation Systems}, 2023.
\bibitem{vaswani2017attention} A. Vaswani, N. Shazeer, N. Parmar, J. Uszkoreit, L. Jones, A. N. Gomez, L. Kaiser, and I. Polosukhin, "Attention is all you need," \textit{Proceedings of the 31st International Conference on Neural Information Processing Systems (NeurIPS)}, pp. 5998–6008, 2017.
\bibitem{liu2021swin} Z. Liu, Y. Lin, Y. Cao, H. Hu, Y. Wei, Z. Zhang, S. Lin, and B. Guo, "Swin transformer: Hierarchical vision transformer using shifted windows," in \textit{Proc. IEEE/CVF Int. Conf. Comput. Vis.}, pp. 10012-10022, 2021.
\bibitem{silberman2012indoor} N. Silberman, D. Hoiem, P. Kohli, and R. Fergus, "Indoor segmentation and support inference from rgbd images," in \textit{Proc. 12th Eur. Conf. Comput. Vis. (ECCV)}, Florence, Italy, pp. 746-760, 2012, Springer.
\bibitem{song2015sun} S. Song, S. P. Lichtenberg, and J. Xiao, "SUN RGB-D: A RGB-D scene understanding benchmark suite," in \textit{Proc. IEEE Conf. Comput. Vis. Pattern Recognit.}, pp. 567-576, 2015.
\bibitem{kingma2014adam} D. P. Kingma and J. Ba, "Adam: A method for stochastic optimization," \textit{Proceedings of the International Conference on Learning Representations (ICLR)}, 2015.
\bibitem{he2016deep} K. He, X. Zhang, S. Ren, and J. Sun, "Deep residual learning for image recognition," in \textit{Proc. IEEE Conf. Comput. Vis. Pattern Recognit.}, pp. 770-778, 2016.
\bibitem{hu2019acnet} X. Hu, K. Yang, L. Fei, and K. Wang, "ACNet: Attention based network to exploit complementary features for RGBD semantic segmentation," in \textit{Proc. IEEE Int. Conf. Image Process. (ICIP)}, pp. 1440-1444, 2019, IEEE.
\bibitem{chen2020bi} X. Chen, K.-Y. Lin, J. Wang, W. Wu, C. Qian, H. Li, and G. Zeng, "Bi-directional cross-modality feature propagation with separation-and-aggregation gate for RGB-D semantic segmentation," in \textit{Eur. Conf. Comput. Vis. (ECCV)}, pp. 561-577, 2020, Springer.
\bibitem{wang2020deep} Y. Wang, W. Huang, F. Sun, T. Xu, Y. Rong, and J. Huang, "Deep multimodal fusion by channel exchanging," \textit{Adv. Neural Inf. Process. Syst.}, vol. 33, pp. 4835-4845, 2020.
\bibitem{yue2021two} Y. Yue, W. Zhou, J. Lei, and L. Yu, "Two-stage cascaded decoder for semantic segmentation of RGB-D images," \textit{IEEE Signal Processing Letters}, vol. 28, pp. 1115-1119, 2021.
\bibitem{zhang2023delivering} J. Zhang, R. Liu, H. Shi, K. Yang, S. Rei{\ss}ner, K. Peng, H. Fu, K. Wang, and R. Stiefelhagen, "Delivering arbitrary-modal semantic segmentation," in \textit{Proc. IEEE/CVF Conf. Comput. Vis. Pattern Recognit.}, pp. 1136-1147, 2023.
\bibitem{pang2023comptr} Y. Pang, X. Zhao, L. Zhang, and H. Lu, "ComPtr: Towards Diverse Bi-source Dense Prediction Tasks via A Simple yet General Complementary Transformer," \textit{arXiv preprint arXiv:2307.12349}, 2023.
\bibitem{girdhar2022omnivore} R. Girdhar, M. Singh, N. Ravi, L. Van Der Maaten, A. Joulin, and I. Misra, "Omnivore: A single model for many visual modalities," in \textit{Proc. IEEE/CVF Conf. Comput. Vis. Pattern Recognit.}, pp. 16102-16112, 2022.
\bibitem{zhang2024lightweight} Y. Zhang, W. Zhou, X. Ran, and M. Fang, "Lightweight Dual Stream Network With Knowledge Distillation for RGB-D Scene Parsing," \textit{IEEE Signal Processing Letters}, vol. 31, pp. 855-859, 2024.
\bibitem{dai2021attentional} Y. Dai, F. Gieseke, S. Oehmcke, Y. Wu, and K. Barnard, "Attentional feature fusion," in \textit{Proc. IEEE/CVF Winter Conf. Appl. Comput. Vis.}, pp. 3560-3569, 2021.
\bibitem{xie2021segformer} E. Xie, W. Wang, Z. Yu, A. Anandkumar, J. M. Alvarez, and P. Luo, "SegFormer: Simple and efficient design for semantic segmentation with transformers," \textit{Adv. Neural Inf. Process. Syst.}, vol. 34, pp. 12077-12090, 2021.
\bibitem{li2023dctnet} J. Li, L. Liu, H. Song, Y. Huang, J. Jiang, and J. Yang, "DCTNet: A Heterogeneous Dual-Branch Multi-Cascade Network for Infrared and Visible Image Fusion," \textit{IEEE Transactions on Instrumentation and Measurement}, 2023.
\bibitem{yang2024min} Y. Yang, M. Wang, S. Huang, and W. Wan, "MIN-MEF: Multi-scale Interaction Network for Multi-exposure Image Fusion," \textit{IEEE Transactions on Instrumentation and Measurement}, 2024.
\bibitem{du2024asymformer} S. Du, W. Wang, R. Guo, R. Wang, and S. Tang, "Asymformer: Asymmetrical cross-modal representation learning for mobile platform real-time rgb-d semantic segmentation," in \textit{Proc. IEEE/CVF Conf. Comput. Vis. Pattern Recognit.}, pp. 7608-7615, 2024.
\bibitem{yang2023pixel} J. Yang, L. Bai, Y. Sun, C. Tian, M. Mao, and G. Wang, "Pixel difference convolutional network for rgb-d semantic segmentation," \textit{IEEE Transactions on Circuits and Systems for Video Technology}, 2023.
\bibitem{bai2022dcanet} L. Bai, J. Yang, C. Tian, Y. Sun, M. Mao, Y. Xu, and W. Xu, "DCANet: differential convolution attention network for RGB-D semantic segmentation," \textit{arXiv preprint arXiv:2210.06747}, 2022.

\bibitem{wang2022rtformer} J. Wang, C. Gou, Q. Wu, H. Feng, J. Han, E. Ding, and J. Wang, "RTFormer: Efficient design for real-time semantic segmentation with transformer," \textit{Advances in Neural Information Processing Systems}, vol. 35, pp. 7423--7436, 2022.

\bibitem{qin2024lkformer} F. Qin, K. Yan, C. Wang, R. Ge, Y. Peng, and K. Zhang, "LKFormer: large kernel transformer for infrared image super-resolution," \textit{Multimedia Tools and Applications}, pp. 1--15, 2024.
\bibitem{ding2022scaling} X. Ding, X. Zhang, J. Han, and G. Ding, "Scaling up your kernels to 31x31: Revisiting large kernel design in CNNs," in \textit{Proceedings of the IEEE/CVF Conference on Computer Vision and Pattern Recognition}, pp. 11963--11975, 2022.
\bibitem{li2024lkca} C. Li, B. Zeng, Y. Lu, P. Shi, Q. Chen, J. Liu, and L. Zhu, "LKCA: Large Kernel Convolutional Attention," \textit{arXiv preprint arXiv:2401.05738}, 2024.

\bibitem{howard2017mobilenets} A. G. Howard, "MobileNets: Efficient convolutional neural networks for mobile vision applications," \textit{arXiv preprint arXiv:1704.04861}, 2017.
\bibitem{chollet2017xception} F. Chollet, "Xception: Deep learning with depthwise separable convolutions," in \textit{Proceedings of the IEEE Conference on Computer Vision and Pattern Recognition}, pp. 1251--1258, 2017.
\bibitem{yu2023metaformer} W. Yu, C. Si, P. Zhou, M. Luo, Y. Zhou, J. Feng, S. Yan, and X. Wang, "Metaformer baselines for vision," \textit{IEEE Transactions on Pattern Analysis and Machine Intelligence}, 2023.

\bibitem{dong2022gebnet} S. Dong, W. Zhou, X. Qian, and L. Yu, "GEBNet: Graph-enhancement branch network for RGB-T scene parsing," \textit{IEEE Signal Processing Letters}, vol. 29, pp. 2273--2277, 2022.

\bibitem{chen2024edge} G. Chen, Q. Wang, B. Dong, R. Ma, N. Liu, H. Fu, and Y. Xia, "EM-Trans: Edge-Aware Multimodal Transformer for RGB-D Salient Object Detection," \textit{IEEE Transactions on Neural Networks and Learning Systems}, 2024.
\bibitem{lee2023decomposed} P. Lee, T. Kim, M. Shim, D. Wee, and H. Byun, "Decomposed Cross-Modal Distillation for RGB-Based Temporal Action Detection," in \textit{Proceedings of the IEEE/CVF Conference on Computer Vision and Pattern Recognition}, 2023, pp. 2373–2383.
\end{thebibliography}
\end{document}


\title{Supplementary Material}

\author{Shuobin Wei, Zhuang Zhou, Zhengan Lu, Zizhao Yuan, and Binghua Su }

\markboth{Journal of \LaTeX\ Class Files, Vol. 14, No. 8, August 2015}
{Shell \MakeLowercase{\textit{et al.}}: Bare Demo of IEEEtran.cls for IEEE Journals}
\maketitle

\begin{abstract}
In the Supplementary Material, we add ablation experiments with HDBFormer. First, we replace MIIM with other interaction modules, finding that MIIM significantly outperforms them in key metrics. Next, we test two feature fusion frameworks, confirming that our chosen dual-branch hierarchical framework is more effective. Finally, we analyze the components of MIIM, showing that each step—particularly the pooling operation—plays a crucial role in reducing computational costs while maintaining performance. These experiments validate our model’s design choices, confirming their effectiveness and efficiency.
\end{abstract}
\IEEEpeerreviewmaketitle

\begin{figure}
\centerline{\includegraphics[width=\columnwidth]{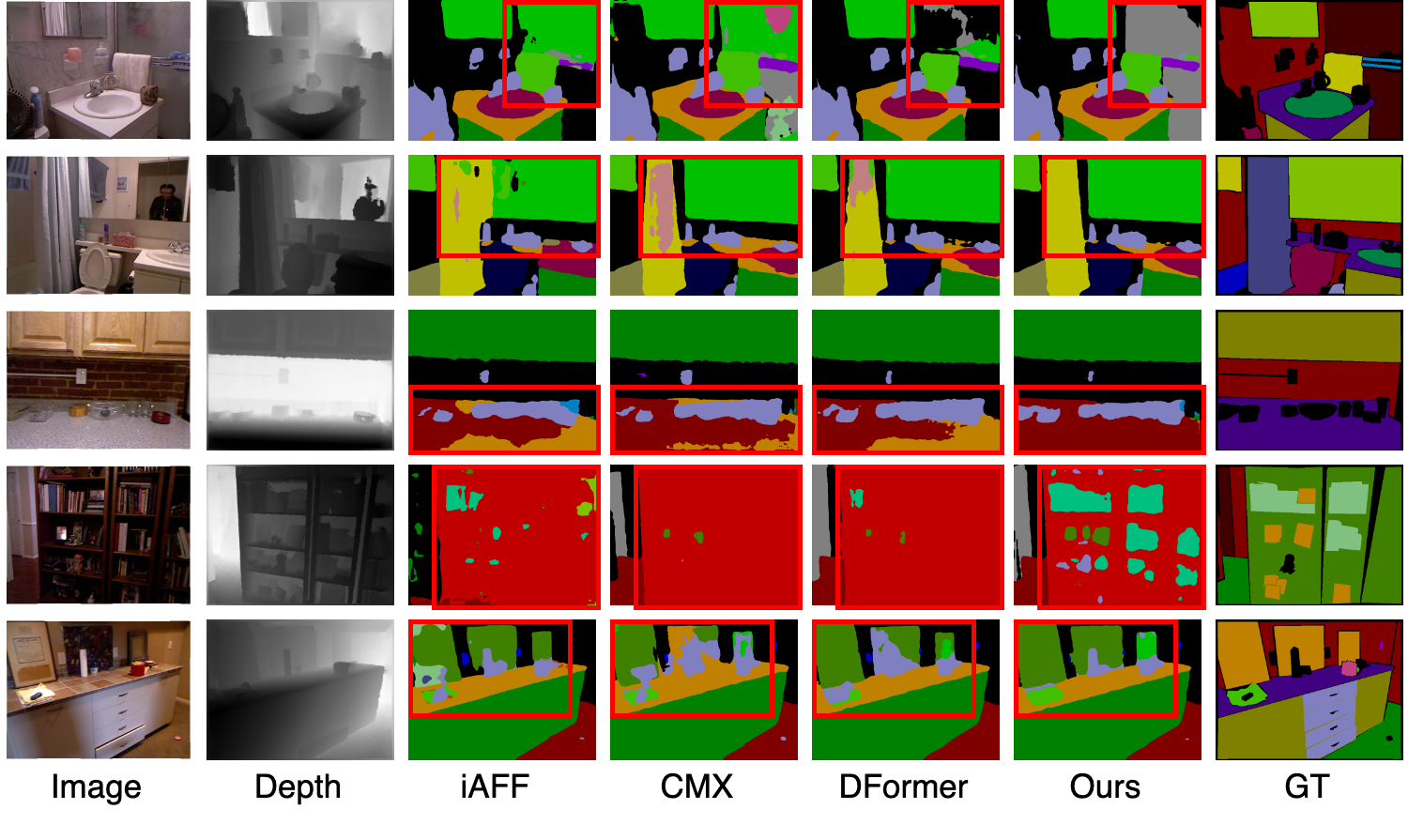}}
\caption{Qualitative analysis of MIIM with other popular modality information interaction module.}
\label{Fig:1}
\end{figure}

\noindent\textbf{1) Ablation experiments with substitution of information interaction modules.}
We conduct comparative experiments between MIIM and other popular modality information interaction module, including pure CNN models and combinations of CNN and Transformer. As shown in \cref{Tab:1}, although iAFF \cite{dai2021attentional} has lower parameters and FLOPs than MIIM, its purely CNN-based structure limits its feature extraction capabilities, resulting in a much lower mIoU compared to MIIM. Therefore, MIIM significantly outperforms existing modality information interaction module in terms of mIoU, parameters, and FLOPs, demonstrating its high efficiency. Additionally, we perform a qualitative analysis of MIIM, as illustrated in \cref{Fig:1}, where MIIM shows a superior ability to leverage the spatial information in depth images to analyze complex RGB scenes, achieving more efficient segmentation.  \vspace{10pt}

\noindent\textbf{2) Ablation experiments with modalities fusion frameworks.}
The purpose of this ablation experiment is to verify the correctness of our chosen feature fusion framework, in which we compare two different feature fusion methods described in Section 1 of this paper. Where framework (a) refers to a two-branch hierarchical architecture and framework (b) refers to a two-branch dual-stream architecture. And in both frameworks, depth separable convolution is used for the basic encoder and Swin Transformer is used for the detail encoder. The results are shown in \cref{Tab:2}, and the experiments demonstrate the effectiveness of our chosen fusion frameworks. \vspace{10pt}

\noindent\textbf{3) Ablation experiments with components of MIIM.}
This ablation experiment aims to verify the effectiveness of each operational step within the MIIM module. Specifically, we remove the pooling operation on the query (Q) in the GFA module and eliminate the concatenation of input features. The experimental results are shown in \cref{Tab:3}, where the $Max$ $Memory$ metric represents the maximum GPU memory usage during model operation. The ablation results confirm that each design step in the MIIM module is reasonable and effective. Notably, removing the pooling operation led to GPU memory usage exceeding the card’s maximum capacity, making it impossible to compute the mIoU under this condition.

\begin{table}[H]
\caption{Ablation experiments with other information interaction module replacing the MIIM module on the NYUDepthv2 dataset.}
\label{Tab:1}
\setlength{\tabcolsep}{5pt}
\begin{tabular}{llccc}
\hline
Fusion Module & Method            & mIoU & \multicolumn{1}{l}{FLOPs(G)} & Params(M) \\ \hline
iAFF \cite{dai2021attentional}          & CNN               & 56.1 & 2.4                          & 2.8       \\
DFormer\cite{yin2023dformer}        & CNN + Transformer & 57.7 & 12.1                         & 23.7      \\
CMX \cite{zhang2023cmx}           & CNN + Transformer & 58.5 & 21.5                         & 14.4      \\ \hline
MIIM (Ours)   & CNN + Transformer &  \textbf{59.3} & 9.7                          & 10.2      \\ \hline
\end{tabular}
\end{table}

\begin{table}[H]
\caption{Ablation experiments of HDBFormer framework with other frameworks on the NYUDepthv2 dataset.}
\label{Tab:2}
\setlength{\tabcolsep}{8pt}
\begin{tabular}{llc}
\hline
Framework              & Constitute           & \multicolumn{1}{c}{mIoU} \\ \hline
{(a)} Two-branch Hierarchical  & Swin-B + DWConv & 57.9                     \\
{(b)} Two-branch Dual-stream  & Swin-B + DWConv & 56.5                     \\ \hline
HDBFormer (Ours)     & Swin-B + Conv + DWConv & \textbf{59.3}            \\ \hline

\end{tabular}
\end{table}

\begin{table}[H]
\caption{Ablation experiments of operations within the GFA component on the NYUDepthv2 dataset.}
\label{Tab:3}
\setlength{\tabcolsep}{22pt}
\begin{tabular}{lcc}
\hline
Operation    & \multicolumn{1}{c}{Max Memory} & mIoU          \\ \hline
w/o Pooling    & 36.127                         & —             \\
w/o Concatenation  & 2.885                          & 58.2          \\ \hline
MIIM (Ours)  & 2.958                          & \textbf{59.3} \\ \hline
\end{tabular}
\end{table}